\documentclass[11pt,a4paper]{article}
\usepackage[hyperref]{emnlp2020}
\usepackage{times}
\usepackage{latexsym}
\usepackage[frozencache,cachedir=.]{minted}
\usepackage{graphicx}
\usepackage[nointegrals]{wasysym}
\usepackage{amsmath}

\usepackage{microtype}

\aclfinalcopy 
\def\aclpaperid{14} 

\title{Going Beyond T-SNE: Exposing \texttt{whatlies} in Text Embeddings}

\author{Vincent D. Warmerdam \\
  Rasa \\
  Sch\"{o}nhauser Allee 175 \\
  10119 Berlin \\
  \texttt{v.warmerdam@rasa.com} \\\And
  Thomas Kober \\
  Rasa \\
  Sch\"{o}nhauser Allee 175 \\
  10119 Berlin \\
  \texttt{t.kober@rasa.com} \\\And
  Rachael Tatman \\
  Rasa \\
  Sch\"{o}nhauser Allee 175 \\
  10119 Berlin \\
  \texttt{r.tatman@rasa.com}\\}

\date{}

\begin{document}
\maketitle
\begin{abstract}
We introduce \texttt{whatlies}, an open source toolkit for visually inspecting word and sentence embeddings.  The project offers a unified and extensible API with current support for a range of popular embedding backends including spaCy, tfhub, huggingface transformers, gensim, fastText and BytePair embeddings. The package combines a domain specific language for vector arithmetic with visualisation tools that make exploring word embeddings more intuitive and concise. It offers support for many popular dimensionality reduction techniques as well as many interactive visualisations that can either be statically exported or shared via Jupyter notebooks. The project documentation is available from \url{https://rasahq.github.io/whatlies/}.
\end{abstract}

\section{Introduction}

The use of pre-trained word embeddings~\citep{Mikolov_2013b,Pennington_2014} or language model based sentence encoders~\citep{Peters_2018,devlin-etal-2019-bert} has become a ubiquitous part of NLP pipelines and end-user applications in both industry and academia. At the same time, a growing body of work has established that pre-trained embeddings codify the underlying biases of the text corpora they were trained on \citep{Bolukbasi_2016, garg2018word, Brunet2019UnderstandingTO}.
Hence, practitioners need tools to help select which set of embeddings to use for a particular project, detect potential need for debiasing and evaluate the debiased embeddings. Simplified visualisations of the latent semantic space provide an accessible way to achieve this.

\begin{figure}[!htb]
\centering
\includegraphics[scale=0.5]{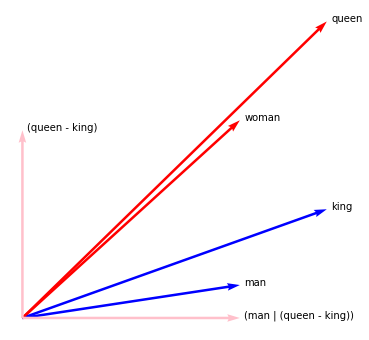}
\captionsetup{font=small}
\caption{Projections of $w_{\text{king}}$, $w_{\text{queen}}$, $w_{\text{man}}$, $w_{\text{queen}} - w_{\text{king}}$ and  $w_{\text{man}}$ projected away from $w_{\text{queen}} - w_{\text{king}}$. Both the vector arithmetic and the visualisation were done using the \texttt{whatlies}. The support for arithmetic expressions is integral in \texttt{whatlies} because it leads to more meaningful visualisations and concise code.}
\label{fig:arithmetic_output}
\end{figure}

Therefore we created \texttt{whatlies}, a toolkit offering a programmatic interface that supports vector arithmetic on a set of embeddings and visualising the space after any operations have been carried out. For example, Figure~\ref{fig:arithmetic_output} shows an example of how representations for \emph{queen}, \emph{king}, \emph{man}, and \emph{woman} can be projected along the axes $v_{\text{queen} - \text{king}}$ and $v_{\text{man} | \text{queen} - \text{king}}$ in order to derive a visualisation of the space along the projections.

The perhaps most widely known tool for visualising embeddings is the tensorflow projector\footnote{\url{https://projector.tensorflow.org/}} which offers 3D visualisations of any input embeddings. The visualisations are useful for understanding the emergence of clusters and the neighbourhood of certain words and the overall space. However, the projector is limited to dimensionality reduction as the sole preprocessing method. More recently,~\citet{molino-etal-2019-parallax} have introduced parallax which allows explicit selection of the axes on which to project a representation. This creates an additional layer of flexibility as these axes can also be derived from arithmetic operations on the embeddings.

The major difference between the tensorflow projector, parallax and \texttt{whatlies} is that the first two provide a non-extensible browser-based interface, whereas \texttt{whatlies} provides a programmatic one. Therefore  \texttt{whatlies} can be more easily extended to any specific practical need and cover individual use-cases. The goal of \texttt{whatlies} is to offer a set of tools that can be used from a Jupyter notebook with a range of visualisation capabilities that goes beyond the commonly used static T-SNE~\citep{Maaten_2008} plots. \texttt{whatlies} can be installed via \texttt{pip}, the code is available from \url{https://github.com/RasaHQ/whatlies}\footnote{Community PRs are greatly appreciated \smiley{}.} and the documentation is hosted at \url{https://rasahq.github.io/whatlies/}.

\section{What lies in \texttt{whatlies} --- Usage and Examples}

\paragraph{Embedding backends.}

The current version of \texttt{whatlies} supports word-level as well as sentence-level embeddings in any human language that is supported by the following libraries:
\begin{itemize}
    \item BytePair embeddings~\citep{sennrich-etal-2016-neural} via the BPemb project~\citep{heinzerling-strube-2018-bpemb}
    \item fastText~\citep{Bojanowski_2017}
    \item gensim~\citep{Rehurek_2010}
    \item huggingface~\citep{Wolf2019HuggingFacesTS}
    \item sense2vec~\citep{Trask_2015}; via spaCy
    \item spaCy\footnote{\url{https://spacy.io/}}
    \item tfhub\footnote{\url{https://www.tensorflow.org/hub}}
\end{itemize}

Embeddings are loaded via a unified API:

\begin{minted}[fontsize=\small]{python}
from whatlies.language import \ 
SpacyLanguage, FasttextLanguage, \ 
TFHubLanguage, HFTransformersLangauge

# spaCy
lang_sp = SpacyLanguage('en_core_web_md')
emb_king = lang_sp["king"]
emb_queen = lang_sp["queen"]

# fastText
ft = 'cc.en.300.bin'
lang_ft = FasttextLanguage(ft)
emb_ft = lang_ft['pizza']

# TF-Hub
tf_hub = 'https://tfhub.dev/google/'
model = tf_hub + 'nnlm-en-dim50/2'
lang_tf = TFHubLanguage(model)
emb_tf = lang_tf['whatlies is awesome']

# Huggingface
bert = 'bert-base-cased'
lang_hf = HFTransformersLanguage(bert)
emb_hf = lang['whatlies rocks']
\end{minted}

In order to retrieve a sentence representation for word-level embeddings such as fastText, \texttt{whatlies} returns the summed representation of the individual word vectors. For pre-trained encoders such as BERT~\citep{devlin-etal-2019-bert} or ConveRT~\citep{Henderson2019ConveRTEA}, \texttt{whatlies} uses its internal \texttt{[CLS]} token for representing a sentence.


\paragraph{Similarity Retrieval.}
The library also supports retrieving similar items on the basis of a number of commonly used distance/similarity metrics such as cosine or Euclidean distance:

\begin{minted}[fontsize=\small]{python}
from whatlies.language import \
SpacyLanguage

lang = SpacyLanguage('en_core_web_md')

lang.score_similar("man", n=5, 
                   metric='cosine')
[(Emb[man], 0.0),
 (Emb[woman], 0.2598254680633545),
 (Emb[guy], 0.29321062564849854),
 (Emb[boy], 0.2954298257827759),
 (Emb[he], 0.3168887495994568)]
 # NB: Results are cosine _distances_
\end{minted}

\paragraph{Vector Arithmetic.}

Support of arithmetic expressions on embeddings is integral in any \texttt{whatlies} functions. For example the code for creating Figure~\ref{fig:arithmetic_output} from the Introduction highlights that it does not make a difference whether the plotting functionality is invoked on an embedding itself or on a representation derived from an arithmetic operation:

\begin{minted}[fontsize=\small]{python}
import matplotlib.pylab as plt
from whatlies import Embedding

man   = Embedding("man", [0.5, 0.1])
woman = Embedding("woman", [0.5, 0.6])
king  = Embedding("king", [0.7, 0.33])
queen = Embedding("queen", [0.7, 0.9])
man.plot(kind="arrow", color="blue")
woman.plot(kind="arrow", color="red")
king.plot(kind="arrow", color="blue")
queen.plot(kind="arrow", color="red")
diff = (queen - king)
orth = (man | (queen - king))

diff.plot(color="pink", 
          show_ops=True)
orth.plot(color="pink", 
          show_ops=True)
# See Figure 1 for the result :)
\end{minted}

This feature allows  users to construct custom queries and use it e.g. in combination with the similarity retrieval functionality. For example, we can validate the widely circulated analogy of~\citet{Mikolov_2013c} on spaCy's medium English model in only 4 lines of code (including imports): $$ w_{\text{queen}} \approx w_{\text{king}} - w_{\text{man}} + w_{\text{woman}}$$

\begin{minted}[fontsize=\small]{python}
from whatlies.language import \
SpacyLanguage

lang = SpacyLanguage('en_core_web_md')

> e = lang["king"] - lang["man"] + \ 
lang["woman"]
> lang.score_similar(e, n=5, 
                     metric='cosine')
[(Emb[king], 0.19757413864135742),
 (Emb[queen], 0.2119154930114746),
 (Emb[prince], 0.35989218950271606),
 (Emb[princes], 0.37914562225341797),
 (Emb[kings], 0.37914562225341797)]
\end{minted}
Excluding the query word \emph{king}\footnote{As appears to be standard practice in word analogy evaluation~\citep{Levy_2014b}.}, the analogy returns the anticipated result: \emph{queen}.

\paragraph{Multilingual Support.} 

\texttt{whatlies} supports any human language that is available from its current list of supported embedding backends. This allows us to check the royal analogy from above in languages other than English. The code snippet below shows the results for Spanish and Dutch, using pre-trained fastText embeddings\footnote{The embeddings are available from \url{https://fasttext.cc/docs/en/crawl-vectors.html.}}.

\begin{minted}[fontsize=\small]{python}
from whatlies.language import \ 
FasttextLanguage
es = FasttextLanguage("cc.es.300.bin")
nl = FasttextLanguage("cc.nl.300.bin")

emb_es = es["rey"] - es["hombre"] + \ 
es["mujer"]
emb_nl = nl["koning"] - nl["man"] + \ 
nl["vrouw"]

es.score_similar(emb_es, n=5,
                 metric='cosine')
[(Emb[rey], 0.04499000310897827), 
(Emb[monarca], 0.24673408269882202), 
(Emb[Rey], 0.2799408435821533), 
(Emb[reina], 0.2993239760398865), 
(Emb[príncipe], 0.3025314211845398)]

nl.score_similar(emb_nl, n=5,
                 metric='cosine')
                 
[(Emb[koning], 0.48337286710739136), 
(Emb[koningen], 0.5858825445175171), 
(Emb[koningin], 0.6115483045578003), 
(Emb[Koning], 0.6155656576156616), 
(Emb[kroonprins], 0.658723771572113)]
\end{minted}

While for Spanish, the correct answer \emph{reina} is only at rank 3 (excluding \emph{rey} from the list), the second ranked \emph{monarca} (female form of \emph{monarch}) is getting close. For Dutch, the correct answer \emph{koningin} is at rank 2, surpassed only by \emph{koningen} (plural of \emph{king}). Another interesting observation is that the cosine distances --- even of the query words --- vary wildly in the embeddings for the two languages. 

\paragraph{Sets of Embeddings.}

In the previous examples we have typically only retrieved single embeddings. However, \texttt{whatlies} also supports the notion of an ``Embedding Set", that can hold any number of embeddings:

\begin{minted}[fontsize=\small]{python}
from whatlies.language import \ 
SpacyLanguage

lang = SpacyLanguage("en_core_web_lg")

words = ["prince", "princess", "nurse", 
         "doctor", "man", "woman",
         "sentences also embed"]
# NB: 'sentences also embed' will be 
#     represented as the sum of the
#.    3 individual words.

emb = lang[words]
\end{minted}

It is often more useful to analyse a set of embeddings at once, rather than many individual ones. Therefore, any arithmetic operations that can be applied to single embeddings, can also be applied to all of the embeddings in a given set. 

The \texttt{emb} variable in the previous code example represents an \texttt{EmbeddingSet}. These are collections of embeddings which can be simpler to analyse than many individual variables. Users can, for example, apply vector arithmetic to the entire \texttt{EmbeddingSet}.

\begin{minted}[fontsize=\small]{python}
new_emb = emb | (emb['man'] - emb['woman'])
\end{minted}


\paragraph{Visualisation Tools.}

Any visualisations in \texttt{whatlies} are most useful when preformed on \texttt{EmbeddingSet}s. They offer a variety of methods for plotting, such as the distance map in Figure~\ref{fig:distance_plot}:

\begin{minted}[fontsize=\small]{python}
words = ['man', 'woman', 'king', 'queen', 
         'red', 'green', 'yellow']
emb = lang[words]
emb.plot_distance(metric='cosine')
\end{minted}

\begin{figure}[!htb]
\centering
\includegraphics[scale=0.5]{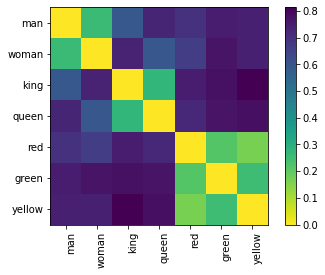}
\captionsetup{font=small}
\caption{Pairwise distances for a set of words using cosine distance.}
\label{fig:distance_plot}
\end{figure}

\begin{figure}[!htb]
\centering
\includegraphics[scale=0.7]{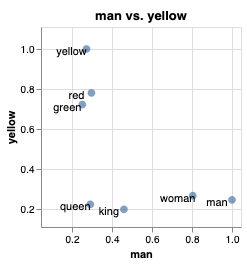}
\captionsetup{font=small}
\caption{Plotting example terms along the axes \emph{man} vs. \emph{yellow}.}
\label{fig:man_yellow}
\end{figure}

\texttt{whatlies} also offers interactive visualisations using ``Altair" as a plotting backend\footnote{Examples of the interactive visualisations can be seen on the project's github page: \url{https://github.com/RasaHQ/whatlies}}:

\begin{minted}[fontsize=\small]{python}
emb.plot_interactive(x_axis="man", 
                     y_axis="yellow",
                     show_axis_point=True)
\end{minted}

The above code snippet projects every vector in the \texttt{EmbeddingSet} onto the vectors on the specified axes. This creates the values we can use for 2D visualisations. For example, given that \emph{man} is on the x-axis the value for `yellow` on that axis will be:

$$v(\text{yellow} \to \text{man}) = \frac{w_{\text{yellow}} \cdot w_{\text{man}}}{w_{\text{man}} \cdot w_{\text{man}}}$$

which results in Figure~\ref{fig:man_yellow}.

These plots are fully interactive. It is possible to click and drag in order to navigate through the embedding space and zoom in and out. These plots can be hosted on a website but they can also be exported to \texttt{png}/\texttt{svg} for publication. It is furthermore possible to apply any vector arithmetic operations for these plots, resulting in Figure~\ref{fig:man_yellow_arithmetic}:

\begin{minted}[fontsize=\small]{python}
e = emb["man"] - emb["woman"]
emb.plot_interactive(x_axis=e, 
                     y_axis="yellow",
                     show_axis_point=True)
\end{minted}

\begin{figure}[!htb]
\centering
\includegraphics[scale=0.7]{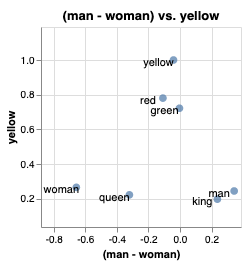}
\captionsetup{font=small}
\caption{Plotting example terms along the transformed \emph{man} - \emph{woman} axis and the \emph{yellow} axis.}
\label{fig:man_yellow_arithmetic}
\end{figure}

\paragraph{Transformations.}

\texttt{whatlies} also supports several techniques for dimensionality reduction of \texttt{EmbeddingSet}s prior to plotting. This is demonstrated in Figure~\ref{fig:pca_umap} below.

\begin{minted}[fontsize=\small]{python}
from whatlies.transformers import Pca
from whatlies.transformers import Umap

p1 = (emb
      .transform(Pca(2))
      .plot_interactive("pca_0", 
                        "pca_1"))
p2 = (emb
      .transform(Umap(2))
      .plot_interactive("umap_0", 
                        "umap_1"))
p1 | p2
\end{minted}

\begin{figure}[!htb]
\centering
\includegraphics[width=\columnwidth]{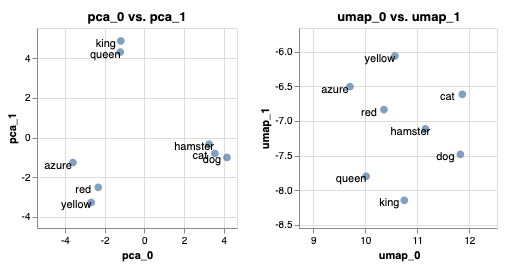}
\captionsetup{font=small}
\caption{Demonstration of PCA and UMAP transformations.}
\label{fig:pca_umap}
\end{figure}

Transformations in \texttt{whatlies} are slightly different than for example scikit-learn transformations because in addition to dimensionality reduction, the transformation can also add embeddings that represent each principal component to the  \texttt{EmbeddingSet} object. As a result, they can be referred to as axes for creating visualisations as seen in Figure \ref{fig:pca_umap}.


\paragraph{Scikit-Learn Integration.}

To facilitate quick exploration of different word embeddings we have also made our library compatible with scikit-learn~\citep{Pedregosa_2011}. The Rasa library uses numpy~\citep{Oliphant_2006} to represent the numerical vectors associated to the input text. This means that it is possible to use the \texttt{whatlies} embedding backends as feature extractors in scikit-learn pipelines, as the code snippet below shows\footnote{Note that this is an illustrative example and we do not recommend to train and test on the same data.}:

\begin{minted}[fontsize=\small]{python}
from whatlies.language import \ 
BytePairLanguage
from sklearn.pipeline import Pipeline

pipe = Pipeline([
    ("embed", BytePairLanguage("en")),
    ("model", LogisticRegression())
])

X = [
    "i really like this post",
    "thanks for that comment",
    "i enjoy this friendly forum",
    "this is a bad post",
    "i dislike this article",
    "this is not well written"
]

y = np.array([1, 1, 1, 0, 0, 0])

pipe.fit(X, y).predict(X)
\end{minted}

This feature enables fast exploration of many different word embedding algorithms.\footnote{At the moment, however, it is not yet possible to use the \texttt{whatlies} embeddings in conjunction with scikit-learn's grid search functionality.} 








    
    


\section{A Tale of two Use-cases}

\paragraph{Visualising Bias.}

\begin{figure}[!htb]
\centering
\includegraphics[scale=0.5]{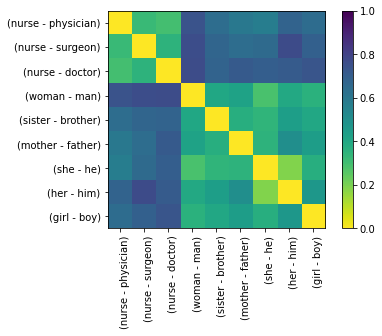}
\captionsetup{font=small}
\caption{Distance map for visualising bias. If there was no bias then we would expect `she-he` to have a distance near 1.0 compared to `nurse-physician`. The figure shows this is not the case.}
\label{fig:biased_embeddings}
\end{figure}

One use-case of \texttt{whatlies} is to gain insight into bias-related issues in an embedding space. Because the library readily supports vector arithmetic it is possible to create an \texttt{EmbeddingSet} holding pairs of representations:

\begin{minted}[fontsize=\small]{python}
lang = SpacyLanguage("en_core_web_lg")

emb_of_pairs = EmbeddingSet(
    (lang["nurse"] - lang["doctor"]),
    (lang["nurse"] - lang["surgeon"]),
    (lang["woman"] - lang["man"]),
)
\end{minted}

Subsequently, the new \texttt{EmbeddingSet} can be visualised as a distance map as in Figure~\ref{fig:biased_embeddings}, revealing a number of spurious correlations that suggest a gender bias in the embedding space.

\begin{minted}[fontsize=\small]{python}
emb_of_pairs.plot_distance(metric="cosine")
\end{minted}

Visualising issues in the embedding space like this creates an effictive way to communicate potential risks of using embeddings in production to non-technical stakeholders. 

It is possible to apply the debiasing technique introduced by~\citet{Bolukbasi_2016} in order to approximately remove the direction corresponding to gender. The code snippet below achieves this by, again, using the arithmetic notation.

\begin{minted}[fontsize=\small]{python}
lang = SpacyLanguage("en_core_web_lg")

emb = lang[words]
axis = EmbeddingSet(
    (lang['man'] - lang['woman']),
    (lang['king'] - lang['queen']),
    (lang['father'] - lang['mother'])
).average()
emb_debias = emb | axis
\end{minted}

\begin{figure}[!htb]
\centering
\includegraphics[scale=0.5]{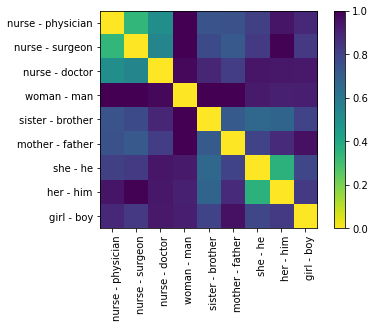}
\captionsetup{font=small}
\caption{Distance map for visualising the embedding space after the debiasing technique of~\citet{Bolukbasi_2016} has been applied.}
\label{fig:debiased_embeddings}
\end{figure}

Figure~\ref{fig:debiased_embeddings} shows the result of applying the debiasing technique, highlighting that some of the spurious correlations have indeed been removed. 

It is important to note though, that the above technique does not reliably remove all relevant bias in the embeddings and that bias is still measurably existing in the embedding space as~\citet{gonen-goldberg-2019-lipstick} have shown. This can be verified with \texttt{whatlies}, by plotting the neighbours of the biased and debiased space:

\begin{minted}[fontsize=\small]{python}
emb.score_similar("maid", n=7)

[(Emb[maid], 0.0),
 (Emb[maids], 0.18290925025939941),
 (Emb[housekeeper], 0.2200336456298828),
 (Emb[maidservant], 0.3770867586135864),
 (Emb[butler], 0.3822709918022156),
 (Emb[mistress], 0.3967094421386719),
 (Emb[servant], 0.40112364292144775)]
 
 emb_debias.score_similar("maid", n=7)

[(Emb[maid], 0.0),
 (Emb[maids], 0.18163418769836426),
 (Emb[housekeeper], 0.21881639957427979),
 (Emb[butler], 0.3642127513885498),
 (Emb[maidservant], 0.3768376111984253),
 (Emb[servant], 0.382546067237854),
 (Emb[mistress], 0.3955296277999878)]

\end{minted}

As the output shows, the neighbourhoods of \emph{maid} in the biased and debiased space are almost equivalent, with e.g. \emph{mistress} still appearing relatively high-up the nearest neighbours list.


\paragraph{Comparing Embedding Backends.}

\begin{figure}[!htb]
\centering
\includegraphics[width=\columnwidth]{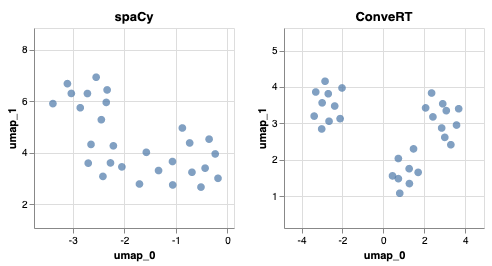}
\captionsetup{font=small}
\caption{Side-by-side comparison of spaCy and ConveRT for embedding example sentences from 3 different intent classes. ConveRT embeds the sentences into relatively tight and coherent clusters, whereas class boundaries are more difficult to see with spaCy.}
\label{fig:spacy_vs_convert}
\end{figure}

Another use-case for \texttt{whatlies} is for comparing different embeddings. For example, we wanted to analyse two different encoders for their ability to capture the intent of user utterances in a task-based dialogue system. We compared spaCy and ConveRT for their ability to embed sentences from the same intent class close together in space. Figure~\ref{fig:spacy_vs_convert} shows that the utterances encoded with ConveRT form tighter and coherent clusters.

\begin{figure*}[!htb]
\centering
\includegraphics[width=\textwidth]{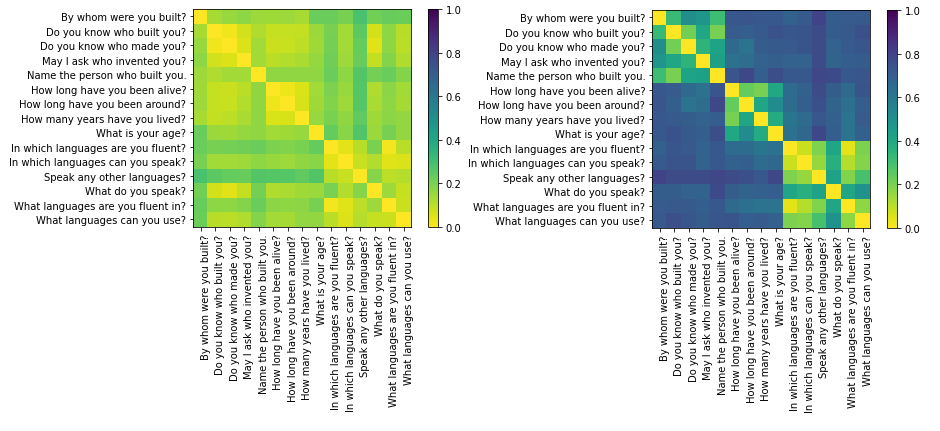}
\captionsetup{font=small}
\caption{Side-by-side comparison of spaCy and ConveRT for embedding example sentences from 3 different intent classes. The distance map highlights the ``clustery" behaviour of ConveRT, where class membership is nicely reflected in the intra-class distances. For spaCy on the other hand, there is less difference between intra-class vs. inter-class distances.}
\label{fig:spacy_vs_convert_dist}
\end{figure*}

Figure~\ref{fig:spacy_vs_convert_dist} highlights the same trend with a distance map, where for spaCy there is barely any similarity between the utterances, the coherent clusters from Figure~\ref{fig:spacy_vs_convert} are well reflected in the distance map for ConveRT. 

The superiority of ConveRT in comparison to spaCy for this example is expected, though, as ConveRT is aimed at dialogue, but it is certainly useful to have a tool --- \texttt{whatlies} --- at one's disposal with which it is possible to quickly validate this.

\section{Roadmap}




\texttt{whatlies} is in active development. While we cannot predict the contents of future community PRs, this is our current roadmap for future development:

\begin{itemize}

    \item We want to make it easier for people to research bias in word embeddings. We will continue to investigate if there are visualisation techniques that can help spot issues and we aim to make any robust debiasing techniques available in \texttt{whatlies}. 
    \item We would like to curate labelled sets of word lists for attempting to quantify the amount of bias in a given embedding space. Properly labelled word lists can be useful for algorithmic bias research but it might also help understand clusters. We plan to make any evaluation resources available via this package. 
    \item One limit of using Altair as a visualisation library is that we cannot offer interactive visualisations with many thousands of data points. We might explore other visualisation tools for this library as well. 
    \item Since we're supporting dynamic backends like BERT at the sentence level, we are aiming to also support these encoders at the word level, which requires us to specify an API for retrieving contextualised word representations within \texttt{whatlies}. We are currently exploring various ways for exposing this feature and are working with a notation that uses square brackets that can select an embedding from the context of the sentence that it resides in:

\begin{minted}[fontsize=\small]{python}
mod_name = "en_trf_robertabase_lg"
lang = SpacyLanguage(mod_name)
emb1 = lang['[bank] of the river']
emb2 = lang['money on the [bank]']
assert emb1.vector != emb2.vector
\end{minted}

    At the moment we only support spaCy backends with this notation but we plan to explore this further with other embedding backends.\footnote{Ideally we also introduce the necessary notation for retrieving the contextualised embedding from a particular layer, e.g. \mintinline{python}{lang['bank'][2]} for obtaining the representation of \emph{bank} from the second layer of the given language model.}

\end{itemize}

\section{Conclusion}

We have introduced \texttt{whatlies}, a python library for inspecting word and sentence embeddings that is very flexible due to offering a programmable interface. We currently support a variety of embedding models, including fastText, spaCy, BERT, or ConveRT. This paper has showcased its current use as well as plans for future development. The project is hosted at \url{https://github.com/RasaHQ/whatlies} and we are happy to receive community contributions that extend and improve the package.

\section*{Acknowledgements}

Despite being only a few months old the project has started getting traction on github and has attracted the help of outside contributions. In particular we'd like to thank Masoud Kazemi for many contributions to the project. 

We would furthermore like to thank Adam Lopez for many rounds of discussion that considerably improved the paper.

\bibliographystyle{acl_natbib}
\bibliography{bib}


\end{document}